\documentclass[letterpaper, 10 pt, conference]{ieeeconf}  % Comment this line out
\IEEEoverridecommandlockouts                              % This command is only
\usepackage[utf8]{inputenc}
\usepackage[T1]{fontenc}
\usepackage{siunitx}
\usepackage{wrapfig,lipsum,booktabs}
\usepackage{url}
\usepackage{graphics} % for pdf, bitmapped graphics files
\usepackage{epsfig} % for postscript graphics files
\usepackage{amsmath} % assumes amsmath package installed
\usepackage{amssymb}  % assumes amsmath package installed
\usepackage{algorithm}% http://ctan.org/pkg/algorithms
\usepackage{algpseudocode}%

\let\bs\boldsymbol
\DeclareMathOperator*{\argmax}{argmax}
\DeclareMathOperator*{\argmin}{argmin}

\newtheorem{remark}{Remark}

\title{\LARGE \bf
CAGES: Cost-Aware Gradient Entropy Search for Efficient Local Multi-Fidelity Bayesian Optimization}

\author{Wei-Ting Tang and Joel A. Paulson% <-this % stops a space
\thanks{W. Tang and J.A. Paulson are with the Dept. of Chemical and Biomolecular Engineering, The Ohio State University, Columbus, OH 43210.
}
\thanks{This work was supported by NSF Grant 2237616.}% <-this % stops a space%
}

\begin{document}

\maketitle
\thispagestyle{empty}
\pagestyle{empty}

%%%%%%%%%%%%%%%%%%%%%%%%%%%%%%%%%%%%%%%%%%%%%%%%%%%%%%%%%%%%%%%%%%%%%%%%%%%%%%%%
\begin{abstract}
Bayesian optimization (BO) is a popular approach for optimizing expensive-to-evaluate black-box objective functions. An important challenge in BO is its application to high-dimensional search spaces due in large part to the curse of dimensionality. 
%that makes it difficult to define a suitable surrogate model needed for optimal sample selection. 
One way to overcome this challenge is to focus on local BO methods that aim to efficiently learn gradients, which have shown strong empirical performance on high-dimensional problems including policy search in reinforcement learning (RL). Current local BO methods assume access to only a single high-fidelity information source whereas, in many problems, one has access to multiple cheaper approximations of the objective. We propose a novel algorithm, Cost-Aware Gradient Entropy Search (CAGES), for local BO of multi-fidelity black-box functions. CAGES makes no assumption about the relationship between different information sources, making it more flexible than other multi-fidelity methods. It also employs a new information-theoretic acquisition function, which enables systematic identification of samples that maximize the information gain about the unknown gradient per evaluation cost. We demonstrate CAGES can achieve significant performance improvements compared to other state-of-the-art methods on synthetic and benchmark RL problems.
\end{abstract}

%%%%%%%%%%%%%%%%%%%%%%%%%%%%%%%%%%%%%%%%%%%%%%%%%%%%%%%%%%%%%%%%%%%%%%%%%%%%%%%%

\section{Introduction}

The problem of optimizing expensive-to-evaluate, noisy black-box functions arises in many real-world applications in science, engineering, machine learning, and beyond. Specific examples include policy search in reinforcement learning (RL) \cite{paulson2023tutorial}, hyperparameter tuning \cite{snoek2012practical}, material design \cite{zhang2020bayesian}, and configuration of high-fidelity physics-based simulators \cite{paulson2022cobalt}. Bayesian optimization (BO) \cite{frazier2018tutorial} is one of the most popular and well studied algorithms for sample-efficient black-box optimization. Although BO has shown good empirical performance on a diverse set of problems, the framework has historically struggled on problems with more than around 10 or so dimensions. This challenge is often referred to as the ``curse of dimensionality'', i.e., BO's cumulative regret scales exponentially with the search space dimension (unless strong assumptions, such as additive structure, are satisfied) \cite{kandasamy2015high}. 

Since global optimization in many dimensions inherently requires more search space exploration, an emerging alternative is to search for locally optimal solutions to high-dimensional objective functions to circumvent this challenge. In fact, this is the same motivation used when training large-scale (deep) neural network architectures, with the key difference being that we cannot directly observe the gradient of the objective function due to its black-box nature. Nonetheless, it is possible to use BO-like approaches to (efficiently) learn the gradient of the objective through noisy observations, which can be used to update the inputs in a way that locally improves the objective. Such recently developed \textit{local} BO methods have shown strong performance on a variety of high-dimensional optimization tasks \cite{muller2021local, nguyen2022local, eriksson2019scalable}. A simple and intuitive example of this type of scheme is \textit{Gradient Information with BO (GIBO)} \cite{muller2021local}, which involves three main steps: (i) construct a Gaussian process (GP) model of the objective function, (ii) use the GP to identify the inputs that are most likely to reduce the average variance of the gradient estimator at a given location, and (iii) combine the gradient estimate with a local optimization algorithm to update the input location. 

In many scenarios, we have access to lower-fidelity approximations of the objective function that can be queried at a cheaper cost than the true objective. A common case where this occurs is when the objective evaluation involves some internal numerical scheme in which one can trade off accuracy for improved computational cost. For example, in policy search for RL, one can increase the integration time, reduce the sample size used to estimate the average reward, and/or replace a detailed physics-based simulator with an approximate version to reduce cost. The problem of integrating these so-called auxiliary ``information sources'' to reduce the cost of BO is often referred to as multi-fidelity BO (MFBO) \cite{kandasamy2016gaussian, poloczek2017multi, takeno2020multi, sorourifar2023computationally}. Many current MFBO methods, however, have been shown to fail when certain underlying assumptions on the auxiliary information sources are not met. A clear example of this behavior is \cite{kandasamy2016gaussian}, which requires the deviation between the true and approximate objective functions to be bounded by a known constant (rarely the case in practice). Additionally, there has been little-to-no work on local MFBO that could address the aforementioned challenges with high-dimensional optimization problems. 

In this work, we propose a local MFBO algorithm called \textit{Cost-Aware Gradient Entropy Search (CAGES)}, which is conceptually simple, efficiently implementable, and generally applicable to multi-fidelity objective functions (i.e., we make no prior assumptions on the relationship between the underlying information sources). CAGES relies on a latent variable Gaussian process (LVGP) model \cite{zhang2020latent}, which utilizes a unique covariance function structure, to enable on-the-fly learning of relationships between the different fidelity levels. It further maximizes a cost-aware acquisition function, which provides a direct measure of improvement in gradient information per query cost, to select the best input-information source pair at every iteration. Gradient information is measured in terms of differential entropy for which we are able to derive a closed-form expression that shows a close relationship to the well-known D-optimality criteria \cite{chaloner1995bayesian} in the design of experiments literature. Through synthetic and real-world functions, we demonstrate (empirically) that our method, CAGES, outperforms both local and global (MF)BO methods and other common baselines. 

The rest of this paper is organized as follows. In Section \ref{sec:preliminaries}, we provide an overview of the relevant background material and the local multi-information source optimization problem of interest. In Section \ref{sec:CAGES}, we derive the proposed CAGES algorithm and discuss some practical implementation details. We evaluate the performance of CAGES on several numerical experiments in Section \ref{sec:numerical-experiments} and provide some concluding remarks in Section \ref{sec:conclusions}. 

\section{Preliminaries}
\label{sec:preliminaries}

\subsection{Bayesian optimization}

Bayesian optimization (BO) aims to globally minimize a black-box function $g : \mathcal{X} \to \mathbb{R}$ in a compact domain $\mathcal{X} \subset \mathbb{R}^d$ of a set of $d$ design (or inputs) parameters, i.e., solving
\begin{align} \label{eq:global-bo}
    \bs{x}^\star \in \argmin_{\bs{x} \in \mathcal{X}} g(\bs{x}),
\end{align}
through possibly noisy queries $y = g(\bs{x}) + \epsilon$ where $\epsilon$ denotes some form of additive noise. BO attempts to solve \eqref{eq:global-bo} by first estimating a \textit{surrogate model} for $g$ from available data, which is used to define a \textit{policy} (specified through maximization of an \textit{acquisition function} defined over the input space) for selecting promising evaluation candidates. The function $g$ is evaluated at the selected candidates (typically at significant expense) and the surrogate model is updated with the newly collected data from which the process can be repeated until the budget is exhausted or a convergence criteria is satisfied. 

\subsection{Gaussian processes (GPs)}
\label{subsec:GPs}

GPs are the most popular class of surrogate models for BO, as they enable flexible, probabilistic non-parametric regression of nonlinear functions \cite{rasmussen2006gaussian, gardner2018gpytorch}. A single-output GP model over an input space $\mathcal{X}$, denoted by $\mathcal{GP}(\mu, k)$, is fully specified by a mean function $\mu : \mathcal{X} \to \mathbb{R}$ and a covariance function $k : \mathcal{X} \times \mathcal{X} \to \mathbb{R}_+$. The covariance function is also called ``the kernel'' due to its connection to kernel methods in the machine learning literature. GPs infer a function $g$ by assuming the output values $g(\bs{X}) \in \mathbb{R}^B$ at any finite collection of inputs $\bs{X} \in \mathbb{R}^{B \times d}$ have a joint Gaussian distribution, i.e., $g(\bs{X}) \sim \mathcal{N}( \mu(\bs{X}), k(\bs{X}, \bs{X}) )$. Standard Gaussian conditioning rules can then be used to condition the GP on a dataset $\mathcal{D}$ (consisting of noisy function observations), which induces an updated posterior GP.
%that captures the information contained in these observations. 
We denote the posterior GP for $g$ conditioned on $\mathcal{D}$ as follows
\begin{align} \label{eq:gp-posterior}
    g \mid \mathcal{D} \sim \mathcal{GP}( \mu_{\mathcal{D}}, k_\mathcal{D} ),
\end{align}
where the posterior mean and covariance are given by
\begin{align*}
    \mu_{\mathcal{D}}(\bs{x}) &= \mu(\bs{x}) + k(\bs{x}, \bs{X}) k(\bs{X}, \bs{X})^{-1} (g(\bs{X}) - \mu(\bs{X})), \\
    k_{\mathcal{D}}(\bs{x}, \bs{x}') &= k(\bs{x}, \bs{x}') - k(\bs{x}, \bs{X})k(\bs{X}, \bs{X})^{-1} k(\bs{X}, \bs{x}),
\end{align*}
and we have overloaded the functions $g(\cdot)$, $\mu(\cdot)$, and $k(\cdot, \cdot)$ to include element-wise operations across their inputs.

\subsection{GP derivatives for local Bayesian optimization}
% \subsection{Multi-information source Bayesian optimization}

An important property of GP models is that they naturally give rise to gradient estimates since GPs are closed under linear operators (such as derivatives) \cite{rasmussen2006gaussian}. Specifically, if $g \sim \mathcal{GP}(\mu, k)$ is a GP with a once-differentiable mean function $\mu$ and a twice-differentiable kernel function $k$, then the \textit{joint} distribution between noisy observations $\bs{y}_{\bs{X}} = g(\bs{X}) + \bs{\epsilon}$ at locations $\bs{X}$ and the gradient at any test point $\bs{x}$ is
\begin{align*}
    &\begin{bmatrix}
        \bs{y}_{\bs{X}} \\
        \nabla g(\bs{x})
    \end{bmatrix} \sim \mathcal{N}\left( \begin{bmatrix}
        \mu(\bs{X}) \\
        \nabla \mu(\bs{x})
    \end{bmatrix}, \begin{bmatrix}
         \tilde{k}(\bs{X}, \bs{X}) & k(\bs{X}, \bs{x}) \nabla^\top \\
        \nabla k(\bs{x}, \bs{X}) & \nabla k(\bs{x}, \bs{x}) \nabla^\top
    \end{bmatrix} \right),
\end{align*}
where $\tilde{k}(\bs{X}, \bs{X}) = k(\bs{X}, \bs{X}) + \sigma^2 \mathbf{I}$ is the covariance function of the noisy samples at evaluation points $\bs{X}$ assuming $\epsilon \sim \mathcal{N}(0, \sigma^2)$ and $\nabla^\top$ operates on the second argument of $k$. This property enables probabilistic inference of the gradient given noisy observations of $g$. The gradient GP conditioned on dataset $\mathcal{D}$ can thus be expressed in terms of \eqref{eq:gp-posterior} as follows
\begin{align} \label{eq:dgp-posterior}
    \nabla g \mid \mathcal{D} \sim \mathcal{GP}( \nabla \mu_{\mathcal{D}}, \nabla k_\mathcal{D} \nabla^\top ).
\end{align}

Local BO methods directly exploit the information in \eqref{eq:dgp-posterior} since, if we can learn any descent direction $\bs{d}_t$ at a point $\bs{x}_t$, then updating it by $\bs{x}_{t+1} = \bs{x}_t + \eta_t \bs{d}_t$ for some step size $\eta_t > 0$ will ensure $g(\bs{x}_{t +1 }) < g(\bs{x}_t)$ (incrementally moving toward our goal of minimizing $g$). 
% $\bs{x}_t$ can be updated to find a point  by $\bs{x}_{t+1} = \bs{x}_t + \eta_t \bs{d}_t$ for $t = 0, 1, \ldots$ where $\eta_t > 0$ is the step size. 
The steepest descent direction corresponds to the negative objective gradient, i.e., $\bs{d}_t = -\nabla g(\bs{x}_t)$ for which we can build a GP model \eqref{eq:dgp-posterior}. The GIBO method \cite{muller2021local} focuses on actively querying samples that minimize the uncertainty in the gradient predictions, as measured by the trace of the posterior covariance matrix. GIBO has been found to achieve promising results on a number of problems, especially as dimensionality $d$ increases. As shown in \cite{wu2024behavior}, GIBO exhibits strong convergence behavior (to local solutions under mild assumptions) and depends only linearly on $d$, which is a significant improvement over standard BO (at the price of potentially not finding the global minimum). 

% \begin{align*}
%     \nabla \mu_D( \bs{x} ) &= \nabla \mu(\bs{x}) - \nabla k(\bs{x}, \bs{X}) \tilde{k}(\bs{X}, \bs{X})^{-1} ( \bs{y} - \mu(\bs{X}) ), \\
%     \nabla k_\mathcal{D} \nabla^\top &= 
% \end{align*}

\subsection{Problem formulation}

Although local BO methods such as GIBO are effective, they require multiple expensive function evaluations at every iteration, which is a limiting factor in many applications. In this work, we consider a variation of \eqref{eq:global-bo} in which one has access to $M$ possibly biased and/or noisy information sources (ISs) for $g$. We denote these ISs by $f^{(\ell)}(\bs{x}) : \mathcal{X} \to \mathbb{R}$ for all $\ell \in \mathbb{N}_1^M \triangleq \{ 1, \ldots, M \}$ and let $f^{(0)} = g$ such that we can observe $g$ directly without bias but possibly with noise. Each IS $\{ f^{(\ell)} \}_{\ell=1}^M$ can be thought of as a ``surrogate'' or ``auxiliary task'' with $\ell = 0$ denoting the ``primary task''. In many real-world applications, such surrogates are readily available or can be derived from simple approximations to the high-fidelity model $g$. Interested readers are referred to \cite{nghiem2023physics} for more information and examples. 

We denote the observations from source $\ell$ at point $\bs{x}$ as
\begin{align}
y_{\bs{x}}^{(\ell)} = f^{(\ell)}(\bs{x}) + \epsilon_{\bs{x}}^{(\ell)}, 
\end{align}
where $\epsilon_{\bs{x}}^{(\ell)} \sim \mathcal{N}(0, \lambda_{\ell}(\bs{x}))$ is an i.i.d. Gaussian noise term with zero mean and variance $\lambda_{\ell}(\bs{x})$ for all $(\bs{x}, \ell) \in \mathcal{X} \times \mathbb{N}_0^M$.
The cost of evaluating task $l$ is given by a function $c_\ell : \mathcal{X} \to \mathbb{R}_{\geq 0}$. For simplicity, we assume that the cost function $c_\ell(\bs{x})$ and the variance function $\lambda_\ell(\bs{x})$ are known and continuously differentiable. In practice, these functions could be estimated from data along with other model parameters (see, e.g., \cite[Chapter 5]{rasmussen2006gaussian} for details). 

In this work, we want to design iterative queries of input-IS pairs $(\bs{x},\ell)$ that maximize gradient information per cost of the query. This can be thought of as an extension of GIBO to handle multi-information source (MIS) objective functions.

\section{Cost-Aware Gradient Entropy Search for Local Multi-Information Source Optimization}
\label{sec:CAGES}

Here, we introduce the CAGES method for locally solving a multi-information source version of \eqref{eq:global-bo}. First, we describe a latent variable GP extension that enables flexible incorporation of data from the ISs with minimal assumptions. Second, we define a cost-aware information-theoretic acquisition function to reduce uncertainty in the primary task gradient. Third, we present the complete CAGES algorithm and discuss some practical implementation choices. 

\subsection{MIS modeling using latent variable GPs (LVGPs)}

The first challenge we encounter in MIS optimization is that we have multiple (potentially correlated) outputs that must be simultaneously modeled in order to fuse information across the ISs. Multi-output Gaussian processes (MOGPs) \cite{liu2018remarks} are a natural extension that assume the outputs follow a multivariate Gaussian distribution. An equivalent MOGP representation can be achieved through the addition of a $(d+1)^\text{th}$ dimension that represents the output index to a single-output GP (Section \ref{subsec:GPs}), which now operates on an augmented space $\mathbb{R}^{d+1}$ through a kernel $k( (\bs{x}, \ell), (\bs{x}', \ell') )$. 

This type of representation has been used in previous works, e.g., \cite{poloczek2017multi}, however, an important question is what kernel structure should be utilized? Standard kernel choices, such as squared exponential (SE), work well for continuous inputs $\bs{x}$ but are not directly applicable to categorical variables $\ell$. 
% A straightforward way to overcome this challenge is to use specific structures, e.g., $k( (\bs{x}, \ell), (\bs{x}', \ell') ) = k_\text{input}(\bs{x}, \bs{x}') \times k_\text{IS}( \ell, \ell' )$; however, they are likely suboptimal since they necessarily impose strong assumptions on how different tasks relate to one another. This is less of a problem in the standard multi-fidelity optimization setting wherein it is assumed that one has strong prior knowledge about the relationship between different tasks 
%(e.g., $\ell$ is defined in a way that the distance $| \ell - \ell' |$ provides an effective measure correlation between the outputs of tasks $\ell$ and $\ell'$). 
% Since we do not assume such knowledge here, we need a more flexible modeling paradigm that learns useful representations from the data. 
We propose the use of latent variable Gaussian processes (LVGPs) to address this challenge, which map the $M+1$ levels of $\ell$ to latent numerical values $\{ \bs{z}(0), \ldots, \bs{z}(M) \}$ \cite{zhang2020latent}. Let $1 \leq m \leq M$ denote the dimension of the latent space. As such, the input $( \bs{x}, \ell )$ is mapped to a new space $( \bs{x}, \bs{z}(\ell) ) \in \mathbb{R}^{d + m}$ for which we can associate a standard kernel such as the SE kernel
\begin{align} \label{eq:mis-kernel}
    & k((\bs{x}, \ell), (\bs{x}', \ell')) \\\notag
    & = \textstyle\zeta^2 \exp\left( -\frac{1}{2}r^2(\bs{x}, \bs{x}') - \| \bs{z}(\ell) - \bs{z}(\ell') \|^2 \right),
\end{align}
where $\zeta^2$ is a scaling factor for the output variance and $r(\bs{x}, \bs{x}') = \sqrt{(\bs{x}-\bs{x}')^\top \Lambda^{-2} (\bs{x}-\bs{x}')}$ is a scaled Euclidean distance with $\Lambda = \text{diag}( l_1, \ldots, l_d )$ denoting a diagonal scaling matrix composed of lengthscale parameters $l_i > 0$. The complete set of hyperparameters that define the LVGP are jointly denoted by $\bs{\xi} = (l_1,\ldots,l_d, \bs{z}(0), \ldots, \bs{z}(M), \zeta^2 ) \in \mathbb{R}^{d + (M+1)m + 1}$. In general, we do not know how to specify $\bs{\xi}$ \textit{a priori} and so look to infer them from data using the maximum likelihood estimation (MLE) framework \cite{zhang2020latent}
\begin{align} \label{eq:mle-opt}
    \bs{\xi}^\star(\mathcal{D}) = \textstyle\argmax_{\bs{\xi}} \mathcal{L}(\bs{\xi} | \mathcal{D}), 
\end{align}
where $\mathcal{L}(\bs{\xi} | \mathcal{D})$ denotes the log-likelihood function under the LVGP model given data $\mathcal{D}$, which has a closed-form expression in terms of covariance matrix obtained by plugging the available samples of $(\bs{x}, \ell)$ into \eqref{eq:mis-kernel}. 
The key takeaway is that the latent variable locations are optimized in LVGPs, enabling them to learn how to most effectively order the ISs. Since only relative distances matter, the first level is always set to the origin in the latent space, i.e., $\bs{z}(0)=\bs{0}$. Although $m$ can be treated as a hyperparameter, the size of $\bs{\xi}$ quickly grows with $m$, increasing the complexity of \eqref{eq:mle-opt}. Thus, we typically set $m=2$ in practice that we found to provide a nice balance between flexibility and complexity. 
% We set $m=2$ that corresponds to a two-dimensional latent space, which has been found to provide better flexibility for capturing correlation between the qualitative factors and helps stabilize the MLE optimization. 
% See the example in Section \ref{subsec:otl-circuit} for an illustration of how such latent representations can be learned from data.

% Since only relative distances matter, the first level $\bs{z}(0)=\bs{0}$ is always set to the origin. Furthermore, to avoid indeterminancy due to rotational/translational invariance, 

% Note that the are no lengthscale parameters for $\bs{z}$ since its scale will be automatically determined in the hyperparameter optimization phase. 
% Under the assumption that the noise variance for each task is known, there are three sets of hyperparameters in the LVGP model: prior variance $\zeta^2$, lengthscales for the design parameters $\bs{\lambda} = (\lambda_1, \ldots, \lambda_d)$, and latent variable values for the task parameter $\bs{Z} = (\bs{z}(1),\ldots,\bs{z}(M))$

\begin{remark}
    There is a close relationship between LVGPs and ``multi-task'' GPs \cite{bonilla2007multi}, which effectively parametrize the kernel $k((\bs{x}, \ell), (\bs{x}', \ell'))$ in a different but related way. In particular, we can rewrite \eqref{eq:mis-kernel} as $K_{i, j} \zeta^2 \exp( -\frac{1}{2}r^2(\bs{x}, \bs{x}') )$ where $K_{i, j} = \exp(-\| \bs{z}(i) -\bs{z}(j) \|^2 )$ for any $i, j \in \{ 0, \ldots, M \}$. A multi-task GP treats the entire positive semi-definite matrix $K$ as a hyperparameter (whose elements are the inter-task similarities). An LVGP, on the other hand, introduces additional structure by assuming the task can be mapped to a lower-dimensional latent space that effectively places a constraint on the relationship between the elements of $K$. 
\end{remark}

\begin{remark}
    A computationally cheaper alternative to using the full LVGP is to replace \eqref{eq:mis-kernel} with the following type of multi-task mixture kernel $(1-\lambda)( k_x(\bs{x},\bs{x}') + k_l( \ell, \ell' ) ) + \lambda k_x(\bs{x},\bs{x}') k_l( \ell, \ell' )$ where $k_x$ is a standard continuous kernel, $k_l$ is a categorical kernel, and $\lambda \in [0,1]$ is a tunable parameter \cite{ru2020bayesian}. The categorical kernel uses an indicator-based similarity metric $k_l( \ell, \ell' ) = \frac{\sigma}{M} \sum_{i=1}^M \mathbb{I}( \ell,\ell' )$ where $\mathbb{I}$ is the indicator function that is 1 if its two arguments are equal and 0 otherwise and $\sigma$ is a scale parameter. This kernel was originally proposed in the context of BO over multiple continuous and categorical inputs and we found it to perform similarly to LVGPs in the case studies presented in Section \ref{sec:numerical-experiments}. This is mainly due to the fact that we do not consider a large number of tasks where LVGPs might uncover more interesting relationships between the tasks. Note that our code implementation includes both types of GPs. 
\end{remark}

\subsection{A cost-aware measure of gradient information}

In BO, acquisition functions measure the expected utility of a sample point given the posterior predictive model. Following the local BO strategy, we want to identify points that are most informative for learning the gradient of $g$ at the current parameters $\bs{x}_t$. Thus, we first propose the gradient entropy search (GES) acquisition that takes an information-theoretic perspective to this problem by characterizing uncertainty in $\nabla g(\bs{x}_t)$ in terms of differential entropy. GES measures the expected reduction in this quantity, i.e., 
\begin{align*}
    \mathrm{H}( \nabla g(\bs{x}_t) | \mathcal{D} ) - \mathbb{E}_{y_{\bs{x}}} \left\lbrace \mathrm{H}( \nabla g(\bs{x}_t) | \mathcal{D} \cup (\bs{x},y_{\bs{x}}) ) \right\rbrace
    % \alpha_\text{GES}(\bs{x}; \bs{x}_t , \mathcal{D}) \\\notag
\end{align*}
where $\mathrm{H}(\bs{Y}) = -\int p(\bs{Y})\log p(\bs{Y}) d \bs{Y}$ is the differential entropy of random vector $\bs{Y} \sim p(\bs{Y})$. Since the gradient GP model \eqref{eq:dgp-posterior} at $\bs{x}_t$ is a multivariate Gaussian, we can derive a closed-form expression for its differential entropy
\begin{align}
    \mathrm{H}( \nabla g(\bs{x}_t) | \mathcal{D} ) = \frac{d}{2}\log(2\pi e | \Sigma'( \bs{x}_t | \mathcal{D}) |^{\frac{1}{d}}),
\end{align}
where $\Sigma'( \bs{x}_t | \mathcal{D}) = \nabla k_\mathcal{D}(\bs{x}_t, \bs{x}_t) \nabla^\top$.
An interesting property of GPs is that the covariance function is independent of the observed target $y_{\bs{x}}$, as shown in \cite{muller2021local}. Thus, the expectation with respect to $y_{\bs{x}} | \mathcal{D}$ can be carried out analytically to yield
\begin{align} \label{eq:ges-simple}
    & \alpha_\text{GES}(\bs{x}; \bs{x}_t, \mathcal{D}) = \textstyle \frac{1}{2}\log| \Sigma'( \bs{x}_t | \bs{X}) | - \frac{1}{2}\log\lvert \Sigma'\left( \bs{x}_t | [\bs{X}, \bs{x}] \right) \rvert, 
\end{align}
where $\bs{X}$ is the set of past data points and $\bs{\hat{X}} = [ \bs{X}, \bs{x} ]$ is a fantasy or virtual dataset that includes the potential future evaluation point $\bs{x}$; the one step ahead covariance matrix depends only on $\bs{\hat{X}}$ due to the aforementioned property. The main difference between GIBO and GES is the operator that is applied to $\Sigma'\left( \bs{x}_t | [\bs{X}, \bs{x}] \right)$ (GIBO is defined by the trace operator $\text{Tr}(\cdot)$ while GES is defined by the determinant $| \cdot |$). This has close connections to the optimal experiment design (OED) literature \cite{chaloner1995bayesian} wherein the trace and determinant operators leads to so-called A- and D-optimality, respectively. GES is thus prioritizing a reduction in the volume of the joint confidence region of the gradient vector as opposed to the average variance of its individual elements. 

Next, we develop a cost-aware version of GES by making two modifications: (i) we replace the standard GP model with an LVGP model, which enables consideration of more than one IS and (ii) we scale the expected information gain by the query cost function, i.e., 
\begin{align} \label{eq:cages}
    & \alpha_\text{CAGES}(\bs{x}, \ell ; \bs{x}_t, \mathcal{D}) = \\\notag
    & \mathbb{E}_{y^{(\ell)}_{\bs{x}}}\left\lbrace \frac{\mathrm{H}( \nabla f^{(0)}(\bs{x}_t) | \mathcal{D} ) - \mathrm{H}( \nabla f^{(0)}(\bs{x}_t) | \mathcal{D} \cup (\bs{x}, \ell, y^{(\ell)}_{\bs{x}}) )}{ c_{\ell}(\bs{x}) } \right\rbrace, \\\notag
    &= \frac{\textstyle \frac{1}{2}\log| \Sigma'( \bs{x}_t, 0 | \bs{X}, \bs{L}) | - \frac{1}{2}\log\lvert \Sigma'\left( \bs{x}_t, 0 | [\bs{X}, \bs{L}, \bs{x}, \ell] \right) \rvert}{c_{\ell}(\bs{x})}.
\end{align}
The numerator is just our GES acquisition for the primary task $\ell=0$ evaluated using the LVGP model, which can be simplified in exactly the same way as done in \eqref{eq:ges-simple}, leading to the final expression in \eqref{eq:cages} where $\bs{L}$ denotes the past set of task levels corresponding to past inputs $\bs{X}$. 
 
% \begin{align}
%     \mathrm{H}( \nabla g(\bs{x}_t) | \mathcal{D} \cup (\bs{x},y_{\bs{x}}) ) = \frac{d}{2}\log(2\pi e \det( \Sigma'( \bs{x}_t | \mathcal{D} \cup ( \bs{x},y_{\bs{x}}) ) )^{1/d})
% \end{align}

% Using the closed-form expression for differential entropy of multivariate Gaussian vectors, we can efficiently maximize GES by
% \begin{align*}
%     \argmax_{\bs{x} \in \mathcal{X}} \alpha_\text{GES}(\bs{x}) &= \argmin_{\bs{x} \in \mathcal{X}} \mathbb{E}\{\log\det( \Sigma'( \bs{x}_t | \mathcal{D} \cup ( \bs{x},y_{\bs{x}}) ) ) \}, \\
%     & = \argmin_{\bs{x} \in \mathcal{X}} \log\det( \Sigma'( \bs{x}_t | [\bs{X}, \bs{x}])),
% \end{align*}

% To account for the MIS setting, we need a cost-aware acquisition function of the form
% \begin{align}
%     \alpha_{\text{CAGES}}()
% \end{align}

\subsection{The CAGES algorithm}

The complete CAGES algorithm is divided into two loops, as shown in Algorithm \ref{alg:cages}. The inner loop selects input-IS pairs that maximize gradient information per query cost and the outer loop updates the current iterate $\bs{x}_t$ using gradient-based optimization. Note that any type of gradient-based optimizer (e.g., Adam or L-BFGS) can be used in place of standard gradient descent in Line 11. The choice of the batch size $B$ is left as a hyperparameter and can be set in multiple ways. It can be adapted at each outer iteration $t$ by not exiting the inner loop until the entropy (or trace of the covariance) is below a threshold or some allotted budget has been exceeded. A simpler choice is to fix $B \sim d$, which is motivated by the fact that $d$ queries to the primary task is enough to exactly learn the gradient in the noiseless setting (assuming the GP hyperparameters are perfectly known) \cite{wu2024behavior}. Although only a heuristic in the MIS setting, we have found it to be effective in the numerical examples presented in the next section. 

\begin{algorithm}[tb!]
\caption{The CAGES algorithm}
\textbf{Input:} black-box MIS functions $\{ f^{(\ell)} \}_{\ell=0}^M$ \\
\textbf{Hyperparameters:} choice of GP kernel and associated hyperpriors, stepsize $\eta$, initial dataset $\mathcal{D}$, number of iterations $T$, and batch size for gradient estimation $B$. \\
\textbf{Initialize:} place LVGP prior on $f^{(\ell)}$ and select initial $\bs{x}_0$. 
\begin{algorithmic}[1]
\For{$t=0, \ldots, T$}
\State Sample primary task: $y_{\bs{x}_t}^{(0)} = f^{(0)}( \bs{x}_t ) + \epsilon^{(0)}_{\bs{x}_t}$.
\State Augment dataset: $\mathcal{D} \leftarrow \mathcal{D} \cup ( \bs{x}_t, 0, y_{\bs{x}_t}^{(0)} )$.
\State Execute GP hyperparameter optimization \eqref{eq:mle-opt}.
\For{$i=1, \ldots, B$}
\State Get new input-IS query point using \eqref{eq:cages}:
\begin{align*}
    ( \bs{x}^\star, \ell^\star ) \in \argmax_{(\bs{x}, \ell) \in \mathcal{X} \times \mathbb{N}_0^M} \alpha_\text{CAGES}( \bs{x}, \ell ; \bs{x}_t, \mathcal{D} ).
\end{align*}
\State Sample task: $y_{\bs{x}^\star}^{(\ell^\star)} = f^{(\ell^\star)}( \bs{x}^\star ) + \epsilon^{(\ell^\star)}_{\bs{x}^\star}$.
\State Augment dataset: $\mathcal{D} \leftarrow \mathcal{D} \cup ( \bs{x}^\star, \ell^\star, y_{\bs{x}^\star}^{(\ell^\star)} )$.
\State Update LVGP posterior for $\nabla f^{(\ell)}( \bs{x}_t )$.
\EndFor
\State Gradient descent: $\bs{x}_{t+1} = \bs{x}_t - \eta \nabla \mu_\mathcal{D}( \bs{x}_t, 0 )$.
\EndFor
\end{algorithmic}
\label{alg:cages}
\end{algorithm}

\section{Numerical Experiments}
\label{sec:numerical-experiments}

In this section, we show results on numerical experiments that compare CAGES to four baseline methods: (1) EI \cite{jones1998efficient}, which is a standard global BO method that maximizes the expected improvement acquisition function at every iteration; (2) GIBO \cite{muller2021local}, which is a single-source local BO method that minimizes the trace of the posterior covariance matrix of the objective gradient; (3) ARS \cite{mania2018simple}, which estimates the objective gradient using finite difference with random perturbations; and (4) MFBO \cite{poloczek2017multi}, which is a global knowledge gradient-based multi-fidelity BO method. Note that since MFBO can query any fidelity level of interest according to the cost-aware acquisition function in \cite{poloczek2017multi}. Therefore, it may never query the high-fidelity function under a limited budget (especially if the query cost is high and the dimension is large). Following standard practice, we minimize the posterior mean function from the GP model to find our best recommended design point at any finite budget value. Also, note that we use the recently proposed logEI \cite{ament2023unexpected} variant of EI that has been recently shown to make the original EI acquisition function much easier to optimize in practice using multi-start gradient-based optimization. 

We estimate the average performance of the algorithms across the randomly drawn initial data $\mathcal{D}$, measurement noise realizations, and random perturbations in ARS by repeating all experiments 10 times from the same random seed. To ensure a fair comparison between the single- and multi-information source BO methods, we allocate a fixed budget for initialization. EI and GIBO only use the initialization budget on the primary task while CAGES and MFBO randomly distribute the budget amongst all tasks. All plots show the mean of the best found objective value as a function of the query budget with the error bars indicating plus/minus one standard error. Our complete implementation is available at: \url{https://github.com/PaulsonLab/CAGES}, which is built upon the BoTorch package \cite{balandat2020botorch}.

\subsection{Rosenbrock benchmark}

The Rosenbrock function is a classic benchmark in the optimization literature. We consider a 12-dimensional version of this problem with two information sources. The primary task $\ell=0$ is the standard Rosenbrock function while the auxiliary task $\ell=1$ includes an oscillatory term \cite{poloczek2017multi}
\begin{align}
    f^{(0)}(\bs{x}) &= \textstyle\sum_{i=1}^{d-1} 100(x_{i+1} - x_i^2)^2 + (x_i - 1)^2, \\\notag
    f^{(1)}(\bs{x}) &= f^{(0)}(\bs{x}) + 0.1 \textstyle\sum_{i=1}^{d-1} \sin(10x_i + 5x_{i+1}), 
\end{align}
where $\bs{x} = (x_1, \ldots, x_{12}) \in \mathbb{R}^{12}$ with domain $\mathcal{X} = [0, 2]^{12}$. We assume a cost of 10 and 1 for each query to $f^{(0)}$ and $f^{(1)}$, respectively. Observations are noise free, though we still treat the noise as a hyperparameter in the GP models.
% We also assume a small amount of noise in both functions, i.e., $\lambda_0(\bs{x}) = \lambda_1(\bs{x}) = 10^{-6}$ for all $\bs{x}$.

\begin{figure}[tb!]
    \centering
    \includegraphics[width=0.5\textwidth]{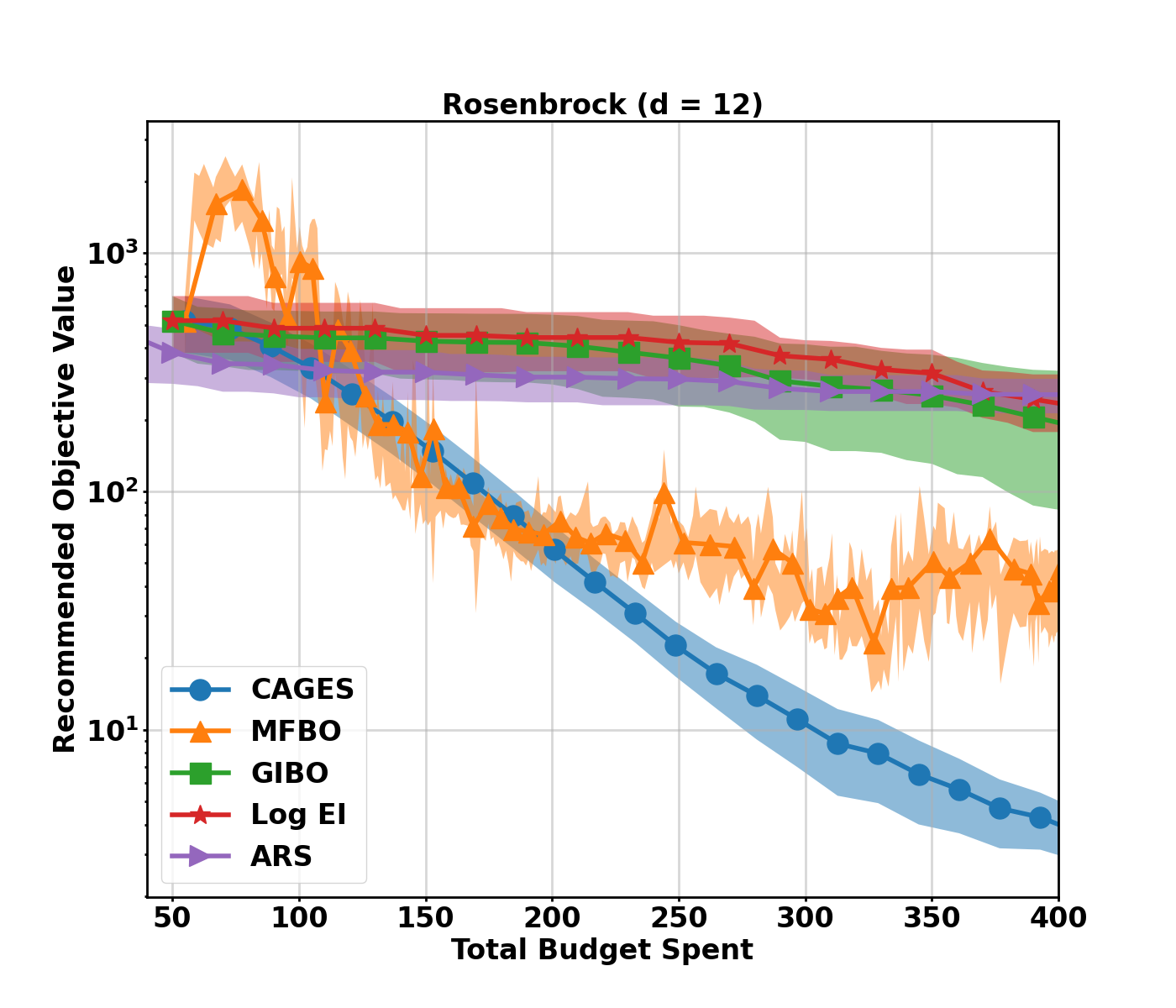}
    \caption{Best found objective value versus total cost for the Rosenbrock benchmark averaged over 10 replicates.}
    % The shaded region are confidence bounds computed using the standard error formula.
    \label{fig:rosenbrock}
\end{figure}

Fig. \ref{fig:rosenbrock} shows the best found objective for each method as a function of the total cost (i.e., cost of initial data plus the accumulated query cost). We see that CAGES significantly outperforms the other methods by achieving lower objectives given the same budget. MFBO is the second-best performing method; however, notice that the objective value for the recommended design has a tendency to fluctuate and even increases for small budgets. This is a consequence of MFBO not necessarily querying the high-fidelity at regular intervals and thus must recommend designs by minimizing the posterior mean GP model. When the model has access to relatively small amounts of data in high dimensions, such recommendations can be inaccurate. CAGES does not exhibit such behavior, as it regularly queries the high-fidelity task after taking a gradient descent step. 
% \footnote{Our goal is to minimize the high-fidelity Rosenbrock function $f^{(0)}(\bs{x})$.}

% higher reward at less total cost. In fact, for all 10 replicates, Rosenbrock found the globally optimal solution using a total budget of less than 250. GIBO is the next best performing method, but finds slightly worse solutions than CAGES while using more than 2x the budget. 
%Interestingly, EI performs the worst on this problem likely due to the cost of additional exploration needed by global BO methods as problem dimension increases.

\subsection{Cartpole-v1}

Next, we focus on a realistic reinforcement learning (RL) problem that we simulate using OpenAI Gym \cite{brockman2016openai}, which provides a suite of environments for testing. We use the same environment for the CartPole-v1 system as in M{\"u}ller et al. \cite{muller2021local}. We use a deterministic neural network control policy that maps 4 states to 2 discrete actions defined in terms of $d=10$ parameters. Additionally, we use the same state and gradient normalization schemes as described \cite{muller2021local} to ensure a fair comparison between algorithms. 

We focus on a total of three information sources, with the primary task $\ell=0$ evaluating the reward over an episode length of 500 steps at an integration time of 0.02 seconds. A reward value of $+1$ is accumulated for every step that the pole remains upright. To ensure the policy is robust to the initial state, we further average the reward across 100 randomly sampled initial states in the domain. An episode ends (terminates) if the pole angle exceeds $\pm 12^\circ$ or the cart leaves the domain. Two cheaper information sources were created by reducing the number of initial states and increasing the integration time. Task $\ell = 1$ considers only 40 of the 100 random initial states and uses an integration time of 0.04, leading to a cost reduction of 5x. Task $\ell = 2$ considers only 10 of the 100 random initial states, leading to a cost reduction of 10x. Therefore, the cost to query $f^{(0)}$, $f^{(1)}$, and $f^{(2)}$ is 10, 2, and 1, respectively, which reflects the true CPU time required for each simulation. 

\begin{figure}[tb!]
    \centering
    \includegraphics[width=0.5\textwidth]{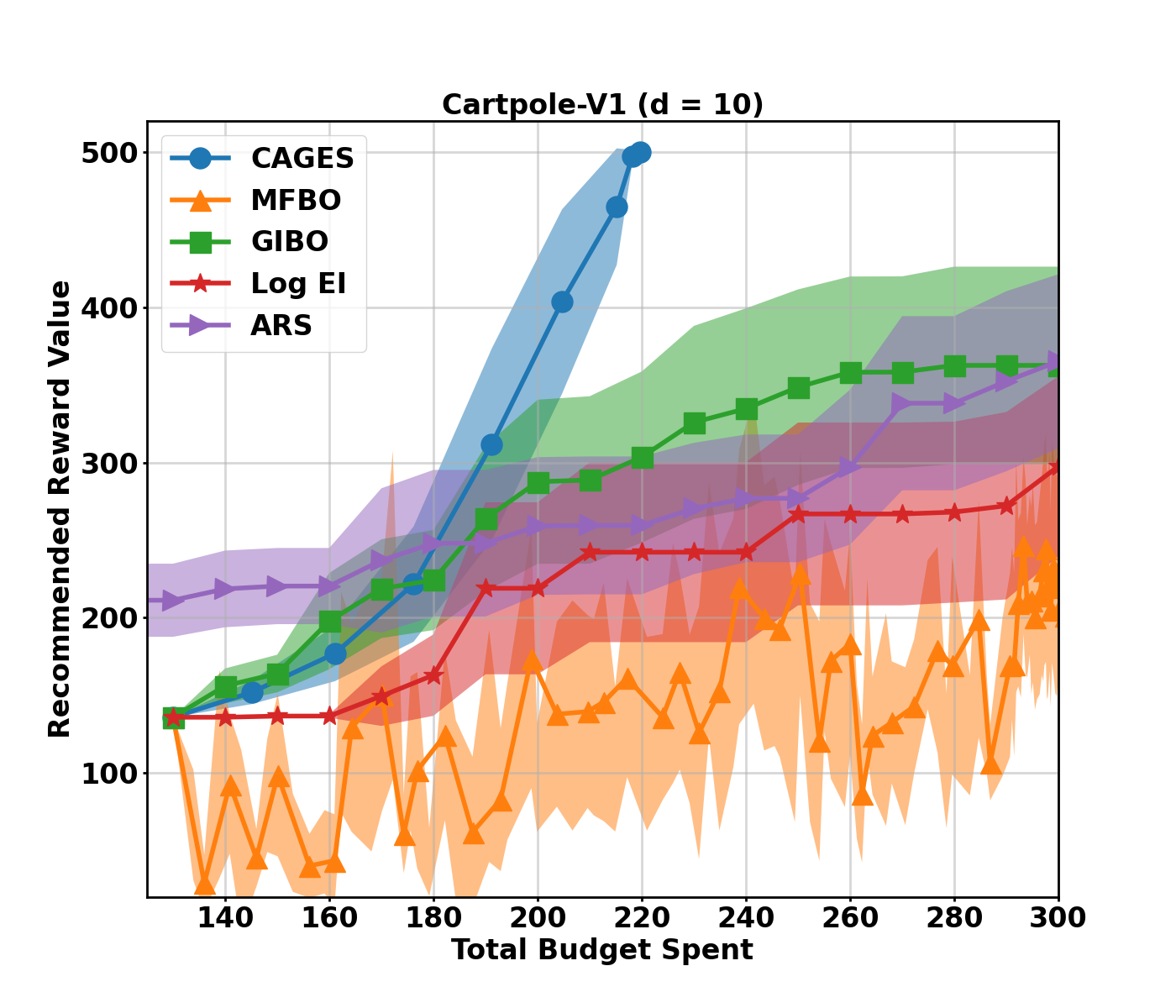}
    \caption{Best found reward value (negative objective) versus total cost for the Cartpole-v1 RL problem averaged over 10 replicates.}
    % The shaded region are confidence bounds computed using the standard error formula.
    \label{fig:cartpole}
\end{figure}

Fig. \ref{fig:cartpole} shows the best found reward value of each method as a function of the total simulation cost. CAGES significantly outperforms the other methods, achieving the maximum possible reward value of 500 for all replicates with a cost of $< 220$. 
% Neither EI or GIBO are able to achieve this level of performance with a cost of $>300$, which highlights the substantial cost savings that can be achieved through a MIS formulation. 
ARS and GIBO are the next best performing methods but only achieve an average reward of $\sim 350$ given a total cost of $300$. The greatly improved performance with CAGES over GIBO highlights the substantial cost savings that can be achieved through a MIS formulation.
Note that logEI, which is a state-of-the-art global BO method, only achieves an average reward of $\sim 300$ given the same budget. MFBO shows significant fluctuations in the reward and only achieves an average reward of $\sim 200$. These latter two results highlight the difficulty of applying global modeling methods in high-dimensional search spaces. Furthermore, although the lower-fidelity tasks clearly provide valuable information for the high-fidelity task, this information cannot be effectively exploited by MFBO in this problem. CAGES, on the other hand, is able to leverage this information locally to achieve significant performance gains. 

It is also worth noting that we did not need to make any assumptions about the relationship between tasks in this problem; it is not obvious if $f^{(1)}$ or $f^{(2)}$ is the better approximation of $f^{(0)}$ but the multi-task GP can learn how to adaptively fuse this information as more data is collected. This problem highlights how combining a local perspective with MIS structure can greatly reduce the resources needed to find high-quality solutions on real-world problems.

% still takes a cost of $>300$ to achieve the same level of performance of CAGES. It is worth noting that one of the biggest strengths of CAGES is that we do not need to make any assumptions about the relationship between the tasks. In this case, it is not obvious if $f^{(1)}$ or $f^{(2)}$ is the better approximation of $f^{(0)}$; however, the LVGP is able to efficiently incorporate both sources of information and adapt on the fly as it learns how to relate these tasks through hyperparameter optimization. 

\section{Conclusions}
\label{sec:conclusions}

In this paper, we develop a local Bayesian optimization algorithm for expensive-to-evaluate, noisy black-box objective functions for which we have access to multiple cheaper approximations of the objective. The proposed algorithm, CAGES, is realized by two key ideas: (1) the use of a latent variable Gaussian process (LVGP) model for flexible multi-task learning from distinct information sources whose underlying relationship is unknown and (2) a mathematically elegant and computationally inexpensive acquisition function that maximizes the gain in gradient information per query cost. We apply CAGES to a synthetic and reinforcement learning (RL) problem where we find that it consistently outperforms known baseline methods. Interesting directions for future work include reducing the computational cost of the LVGP, developing a theoretical analysis of the convergence properties of CAGES, and applying CAGES to even more complicated RL problems. 

% Future work

\bibliographystyle{IEEEtran}
\bibliography{reference}

\end{document}